\documentclass{llncs}
\usepackage{caption}
\usepackage{multibib}
\newcites{languageresource}{Language Resources}
\usepackage{graphicx}
\usepackage{tabularx}
\usepackage{soul}
\usepackage{algorithm,algorithmic}
\usepackage[T1]{fontenc}
\usepackage{textgreek}
\usepackage{multirow}
\usepackage{enumitem}
\usepackage{amsmath}
\usepackage{xcolor}
\usepackage[normalem]{ulem}
\usepackage[utf8]{inputenc}
\useunder{\uline}{\ul}{}

\setlist{noitemsep}

\begin{document}

\title{Analyzing Roles of Classifiers and Code-Mixed factors for Sentiment Identification}
\titlerunning{Hamiltonian Mechanics}  % abbreviated title (for running head)
\author{Soumil Mandal\inst{1}, Dipankar Das\inst{2}}

\institute{Department of Computer Science \& Engineering, SRM University, Chennai\\
\and
Department of Computer Science \& Engineering, Jadavpur University, Kolkata\\
\textcolor{black!50}{\{soumil.mandal, dipankar.dipnil2005\}@gmail.com}}

\maketitle              % typeset the title of the contribution

\begin{abstract}
Multilingual speakers often switch between languages to express themselves on social communication platforms. Sometimes, the original script of the language is preserved, while using a common script for all the languages is quite popular as well due to convenience. On such occasions, multiple languages are being mixed with different rules of grammar, using the same script which makes it a challenging task for natural language processing even in case of accurate sentiment identification. In this paper, we report results of various experiments carried out on movie reviews dataset having this code-mixing property of two languages, English and Bengali, both typed in Roman script. We have tested various machine learning algorithms trained only on English features on our code-mixed data and have achieved the maximum accuracy of 59.00\% using Na\"ive Bayes (NB) model. We have also tested various models trained on code-mixed data, as well as English features and the highest accuracy of 72.50\% was obtained by a Support Vector Machine (SVM) model. Finally, we have analyzed the misclassified snippets and have discussed the challenges needed to be resolved for better accuracy.     
\end{abstract}

\section{Introduction}
English is by far the most popular language in the Web 2.0, but on social media, its dominance is receding. An automated language detection algorithm was applied to over 62 million tweets to identify the top 10 most popular languages on Twitter [1]. It was found that about half of the tweets were in English while the other half were in other languages. It is also a popular trend, and growing with the rise in middle class in several countries to mix multiple languages (code-mixing) for expressing their thoughts on social media. Users whose first language uses non-Roman alphabets type in Roman script for convenience. Such usage increases the likelihood of code-mixing. This case is quite clearly observed in South Asia and especially in the Indian subcontinent. Majority of schools in the urban areas of India use English as the primary language of teaching and communication and this leads to rapid increase in code-mixed data that isn’t being utilized to its potential due to lack of resources and systems which can deal this effectively. For our experiments, we used data comprising of two languages, English (the most popular language for international communication purposes, spoken by 5.52\%  of the world population as of 2010) and Bengali (the dominating language in the region of West Bengal and Bangladesh, spoken by 3.05\% of the world population as of 2010). \\  
\hspace*{0.5cm}Sentiment analysis which is also known as opinion mining is rapidly growing field in the world of Natural Language Processing. In cases such as these where the user mixes several languages, the task of sentiment analysis becomes harder and accuracy decreases rapidly. The task of sentiment classification of multilingual text has been attempted by various researchers before. Most of the works have been carried out on data where the original script of the languages was used, whereas there have been quite a few works where a common script has been used as well. One frequently used method is to convert a whole document into a single language followed by determining it's polarity [2]. This method is not quite accurate due to the fact that machine translation in itself is a big challenge and many a time several classes of information is lost in the process. A method where classifiers trained on individual languages has been explored as well [3]. More complex methods have been tested like language identification followed by POS tagging and finally polarity identification [4]. This process is relatively ineffective in cases like linguistic code-switching where loss of context is a big issue. Experiments on cross-lingual sentiment analysis have been tried as well [5]. On the other hand, a language-independent model, relying only on emoticons which outperformed a Na\"ive Bayes model trained on the bag of words is described in [6]. A method which does not rely only on emoticons, but also character and punctuation repetitions and considers language independent features is demonstrated in [7]. To the best of our knowledge, there hasn’t been any work on sentiment identification of English-Bengali code-mixed data yet where both the languages are in Roman script. Many challenges can be seen in this scenario, the most common ones are present in grammatical structure as well as ambiguity. \\     
\hspace*{0.5cm}In our paper, we aim to see how different supervised learning algorithms trained on English features perform on English-Bengali code-mixed data as well as improve the accuracy of the same by including code-mixed features in the training set. Both the English and code-mixed data are on the same topic in our case, which is movie/film reviews. On a whole, we have performed three experiments. In the first experiment, we have trained our classifiers on English features and have tested the same on English movie reviews dataset where LSVC (SVM with the linear kernel) obtained the best accuracy with a score of 84.45\%. For the second experiment, we have again used the previously trained classifiers and have tested them on our code-mixed dataset. Here, MNB (Multinomial Na\"ive Bayes) was the winner with an accuracy of 59.00\%. Finally, in our third experiment, we have extracted features from our code-mixed training dataset and have tested them on code-mixed data and again, LSVC performed the best with an accuracy of 72.50\%. Lastly, important evaluating parameters of the best performing classifiers from each experiment were calculated and the misclassified snippets were analyzed. We have also discussed the possible steps needed to be taken to improve accuracy in future. \\ 
\hspace*{0.5cm} The paper is organized as follows: In Section~\ref{sec2}, we have described the datasets used for experimentation. In Section~\ref{sec3}, we have described the machine learning algorithms used for our work and in Section~\ref{sec4} we have discussed the features used to train these algorithms. The experimental setup is described in Section~\ref{sec5} along with results obtained. In Section~\ref{sec6}, we have analyzed the misclassified snippets and have discussed the probable reasons. Finally, in Section~\ref{sec7}, we concluded the paper and have discussed the work needed to be done in the future for better results. 

\section{Data Sets}
\label{sec2}
On a whole, two datasets were used for conducting our experiments. One was in English while the other one in English-Bengali code-mixed. As Bengali is a low resource language, especially with respect to social media, the amount of code-mixed data collected was comparatively less.

\subsection{English Data}
The Cornell polarity dataset v1.0~\footnote{
http://www.cs.cornell.edu/people/pabo/movie-review-data/}  consisting of movie reviews has been used in our present experiments. It contained 5331 positive snippets and 5331 negative snippets. Training dataset was made by randomly picking 4000 snippets from each of the two sets. The test dataset contains randomly picked 1200 snippets from the 1331 remaining snippets from each of the sets. The two datasets had no snippet in common. Distribution of data is shown in Table~\ref{table1}.
\begin{table}[H]
\begin{tabular}{|c|c|c|c|}
\hline
\textbf{Data} & \textbf{Positive} & \textbf{Negative} & \textbf{Total} \\ \hline
Training & 4000 & 4000 & 8000 \\ \hline
Testing & 1200 & 1200 & 2400 \\ \hline
\end{tabular}
\centering
\caption{English data distribution.}
\label{table1}
\end{table}

Examples \\
1. A real movie, about real people, that gives us a rare glimpse into a culture most of us don't know. \\
2. Intriguing and beautiful film, but those of you who read the book are likely to be disappointed. \\
3. It was a wonderful day in the beach and I thoroughly enjoyed it.

\subsection{Code-Mixed Data}
Our code-mixed data comprises of two languages, namely English and Bengali. The matrix and embedded languages are not fixed, i.e. in some cases Bengali grammar is used with English words, while in some cases, it's the other way around. The data was collected using the Twitter API~\footnote{https://developer.twitter.com/en/docs}  and the Facebook Graph API~\footnote{https://developers.facebook.com/docs/graph-api/} by querying them with commonly used Bengali words transliterated in Roman, e.g. \textit{bhalo} (meaning good), \textit{baje} (meaning bad), \textit{kharap} (meaning worse). We collected data from June 15, 2016 to Dec 1, 2016 (duration of 5 months). Reviews were handpicked based on relevance to the topic (movie/film/show). A total of 800 positive and 800 negative snippets were collected. For training purpose, 600 positive and 600 negative snippets were selected using a random function. For testing, the remaining 200 positive and 200 negative snippets were used. The data used for training and test had no snippets in common. Our code-mixed data is relatively quite small due to the fact that it’s extremely tedious and time consuming to collect snippets consisting of solely English and Bengali (Roman script) sentences, as none of the mentioned APIs provide such facility where we can specify this. Distribution of data is showed in Table~\ref{table2}. \\
\begin{table}[H]
\centering
\begin{tabular}{|c|c|c|c|}
\hline
\textbf{Data} & \textbf{Positive} & \textbf{Negative} & \textbf{Total} \\ \hline
Training & 600 & 600 & 1200 \\ \hline
Testing & 200 & 200 & 400 \\ \hline
\end{tabular}
\caption{Code-mixed data distribution.}
\label{table2}
\end{table}
Examples \\
1. (Etotai kharap je)\textsubscript{BN} (critics)\textsubscript{EN} (der khub koshto kore dekhte hoyeche)\textsubscript{BN} (film)\textsubscript{EN} (tah)\textsubscript{BN} . \\
\underline{Translation}: So pathetic that the critics had a tough time watching the film. \\
2. (Bondhuder sathe dekhar jonho ekdom thik thak)\textsubscript{BN} (movie)\textsubscript{EN} (tah)\textsubscript{BN} , (Shei)\textsubscript{BN} (school days)\textsubscript{EN} (er kotha mone pore gelo)\textsubscript{BN} . \\
\underline{Translation}: The movie is ideal for watching with friends, reminds me of my school days. \\
3. (Proshongsha chara r kichui nei amar mukhe)\textsubscript{BN} , (bhaggish)\textsubscript{BN} (first day first show)\textsubscript{EN} (tei gechilam nahole ekgada)\textsubscript{BN} (spoilers)\textsubscript{EN}  (shunte hoto)\textsubscript{BN} . \\
\underline{Translation}: I have nothing but appreciation, fortunately watched first day first show otherwise would have had to hear a lot of spoilers. 

\section{Supervised Classifiers Used}
\label{sec3}
In our present set up, we have not implemented any unsupervised system because the identification of sentiment using unsupervised methods has not produced satisfactory results [17]. Moreover, to deal with special code-mixed features, usage of supervised models is necessary. For all of them, the implementation in scikit-learn~\footnote{http://scikit-learn.org/stable/} package was employed keeping parameters at default.

\subsection{Na\"ive Bayes (NB)}
These methods are a set of supervised learning algorithms based on Bayes Theo-rem with the na\"ive assumption of independence between every pair of features. The different Na\"ive Bayes classifiers differ mainly by the assumptions they make regarding prob(x/y). For our experiments, we have used Gaussian Na\"ive Bayes (GNB), Bernoulli Na\"ive Bayes (BNB) and Multinomial Na\"ive Bayes (MNB).

\subsection{Linear Model (LM)}
Linear models describe a continuous response variable as a function of one or more predictor variables. We have used two such models for our experiments. The first one is Logistic Regression (LRC) whereas the second classifier is called Stochastic Gradient Descent (SGDC) which is based on partial fit method. 

\subsection{Support Vector Machine (SVM)}
Support vector machines are supervised learning models with associated learning algorithms that analyze data used for classification and regression analysis. They are large-margin rather than probabilistic classifiers in contrast to Na\"ive Bayes and Linear Models. They are found to be highly effective for text classification and generally outperform Na\"ive Bayes models. For our experiments, we have used Linear Support Vector Machine (LSVC) which is based on \textit{one vs the rest} and NuSVC which is based on \textit{one against one}.

\section{Features}
\label{sec4}
For creating any supervised model, selection of proper features is very important for getting good results. The features which are discussed below.

\subsection{Part Of Speech (POS)}
A part of speech tagger is a piece of software that reads text and assigns part of speech tag to each word based on context. We have used the NLTK POS tagger  for our experiments. Prior to feeding the data to the tagger, we conducted a preprocessing step using NLTK regex tokenizer which removed the punctuation marks. POS tagging is important as different parts of speech posses varying importances, for example in case of opinion mining, adjectives tend to contribute the most to the overall sentiment.

\subsection{N-Grams}
N-Gram refers to contiguous sequence of \textit{n} items from a given sequence of text or speech. For our experiments, we have used unigrams, bigrams and trigrams (n = 1 to 3). N-Grams play an important role in context capture. For example, the word \textit{good} which is a positive word with a negation prior to it, like \textit{not} makes it a negative phrase. We have generated the n-grams with the help of NLTK n-gram module. \\
Example: (Movie)\textsubscript{EN} (ta khub bhalo)\textsubscript{EN} . \\
\underline{Translation}: The movie is very good. \\
unigrams $-$ [\{Movie\}, \{ta\}, \{khub\}, \{bhalo\}], bigrams $-$ [\{Movie ta\}, \{ta khub\}, \{khub bhalo\}], trigrams $-$ [\{Movie ta khub\}, \{ta khub bhalo\}] 

\subsection{SWN 3.0}
A word appearing in the SentiWordNet (SWN) [8] generally contains emotion. For word level emotion classification, it is necessary to disambiguate emotion and non-emotion words properly. This feature helps the classifier to clearly define emotion and non-emotion words. We have used SWN 3.0  present in the NLTK package for our experiments. 

\subsection{SO-CAL}
SO-calculator [9] is an application used for calculating semantic orientation of text documents. It was mainly designed for online product reviews. It has a total of five dictionaries, namely adjective, adverb, noun, verb and intensification.

\subsection{NRC emotion lexicon}
The NRC emotion lexicon  [10] is a list of English words (about 14,000) and their association with eight basic emotions (anger, fear, anticipation, trust, surprise, sadness, joy and disgust) and two sentiments (negative and positive). 

\section{Experimental Setup \& Analysis}
\label{sec5}
On a whole, three experiments were performed on English and code-mixed data. The setup along with the analysis is described in detail below.

\subsection{English Data}
\textbf{Exp1}: For our unigram model, we had done experiments using adjectives only (fe01) as well as adjectives, nouns, adverbs and verbs (fe02). After extraction of n-grams, we discarded bigrams with $\geq$ 1 stop-words and trigrams with $\geq$ 2 stop-words. The top 1000  bigrams (fe03) and top 500 trigrams (fe04) were chosen based on tf-idf. They were used in a bag of words fashion. The lexicons used for extracting features from the training set were SWN (fe05), SOCAL (fe06), and NRC emotion lexicon (fe07). For NRC emotion lexicon, words without a polarity (i.e. positive 0, negative 0) were not used. Features used for training are described in FSet1. Results are shown in Table~\ref{table3}.
\\ \\
Feature set acronyms (FSet1):
\\
\hspace*{0.5cm}fe01 $\rightarrow$ unigrams: all adjectives \\ 
\hspace*{0.5cm}fe02 $\rightarrow$ unigrams: all adjectives, nouns, adverbs, verbs \\
\hspace*{0.5cm}fe03 $\rightarrow$ fe02 + top 1000 bigrams \\
\hspace*{0.5cm}fe04 $\rightarrow$ fe03 + top 500 trigrams \\
\hspace*{0.5cm}fe05 $\rightarrow$ fe04 + swn \\
\hspace*{0.5cm}fe06 $\rightarrow$ fe05 + socal \\
\hspace*{0.5cm}fe07 $\rightarrow$ fe06 + nrc

\begin{table}[H]
\begin{tabular}{|l|c|c|c|c|c|c|c|}
\hline
\multicolumn{1}{|c|}{\multirow{3}{*}{Feature Combinations}} & \multicolumn{7}{c|}{\textbf{Accuracy in \%}} \\ \cline{2-8} 
\multicolumn{1}{|c|}{} & \multicolumn{3}{c|}{NB} & \multicolumn{2}{c|}{LM} & \multicolumn{2}{c|}{SVM} \\ \cline{2-8} 
\multicolumn{1}{|c|}{} & GNB & BNB & MNB & LRC & SGDC & LSVC & NuSVC \\ \hline
uni (adj) & 61.83 & 67.58 & 70.12 & 71.08 & 69.58 & 72.80 & 72.45 \\ \hline
uni (adj + adv + vrb + nou) & 65.58 & 74.41 & 75.12 & 76.00 & 73.29 & 78.16 & 77.79 \\ \hline
uni + bi & 68.04 & 76.41 & 77.41 & 76.62 & 73.95 & 80.45 & 80.37 \\ \hline
uni + bi + tri & 68.70 & 76.54 & 77.58 & 76.79 & 74.12 & 81.04 & 81.08 \\ \hline
uni + bi + tri + swn & 71.12 & 79.00 & 79.91 & 79.50 & 76.41 & 83.75 & 83.62 \\ \hline
uni + bi + tri + swn + socal & 71.62 & 79.25 & 80.33 & 79.91 & 76.70 & 84.20 & 84.08 \\ \hline
uni + bi + tri + swn + socal + nrc & 71.91 & 79.50 & 80.58 & 80.08 & 76.75 & \textbf{84.45} & 84.37 \\ \hline
\end{tabular}
\captionsetup{justification=centering}
\caption{Accuracies of classifiers (English train - English test). Boldface: Best performance.}
\label{table3}
\end{table}

\textbf{Summary}: From Table~\ref{table3}, we can clearly see a jump in the accuracy from fe01 (max
acc. 72.80\%) to fe02 (max acc. 78.16\%). Introducing bigrams fe03 increased the accuracy
for all the classifiers (average 1.65\%) and introducing trigrams fe04 increased the
accuracy as well (average 0.37\%), though not as much as bigrams. Introduction of
SWN (fe05) again shows a significant improvement in the accuracy (average
2.49\%). There on, introduction of the other lexicons, namely SOCAL (fe06) and NRC emotion lexicon (fe07)
showed little improvement in the accuracy. This is probably due to the fair amount of intersection between the lexicons. \\ 
\textbf{Classifiers}: It is clear from Table~\ref{table3} that Support Vector Machines (LSVC and NuSVC)
performed better than the rest [11]. Among these two, LSVC (confusion matrix shown
in Table 7) got the edge with an accuracy of 84.45\% whereas NuSVC got 84.37\%. In
linear models, LRC performed slightly better with an accuracy of 80.08\% while
SGDC performed significantly poorer with an accuracy of 76.75\%. Na\"ive Bayes models
performed comparatively well with MNB getting the highest accuracy of 80.58\%,
BNB with an accuracy of 79.50\% and GNB got the overall least accuracy with a score
of 71.91\%. For our case (English train - English test), based on performance, we
can infer that SVM > NB > LM.

\subsection{Code-Mixed Data}
On code-mixed testing dataset, we have done two experiments, namely English train $-$ code-mixed test (exp2) and code-mixed train $-$ code-mixed test (exp3). \\

\textbf{Exp2}: Here we have ran the same classifiers used in exp1 i.e. classifiers trained on English based on fe01, fe02, fe03, fe04, fe05, fe06 and fe07 (FSet1) features and tested on code-mixed data. The performances of the different classifiers are shown in Table~\ref{table4}.

\begin{table}[H]
\begin{tabular}{|l|c|c|c|c|c|c|c|}
\hline
\multicolumn{1}{|c|}{\multirow{3}{*}{Feature Combinations}} & \multicolumn{7}{c|}{\textbf{Accuracy in \%}} \\ \cline{2-8} 
\multicolumn{1}{|c|}{} & \multicolumn{3}{c|}{NB} & \multicolumn{2}{c|}{LM} & \multicolumn{2}{c|}{SVM} \\ \cline{2-8} 
\multicolumn{1}{|c|}{} & GNB & BNB & MNB & LRC & SGDC & LSVC & NuSVC \\ \hline
uni (adj) & 45.25 & 50.75 & 51.50 & 50.00 & 49.75 & 50.00 & 49.75 \\ \hline
uni (adj + adv + vrb + nou) & 50.00 & 53.75 & 54.50 & {54.25} & {54.25} & {52.50} & 53.50 \\ \hline
uni + bi & 50.75 & 54.25 & 55.25 & {53.75} & {53.50} & {53.00} & 54.25 \\ \hline
uni + bi + tri & 51.00 & 55.25 & 55.50 & {53.00} & {53.25} & {53.00} & 54.75 \\ \hline
uni + bi + tri + swn & 53.75 & 57.50 & 58.25 & 54.50 & 54.50 & 55.25 & 57.00 \\ \hline
uni + bi + tri + swn + socal & 54.25 & 57.75 & 58.25 & 55.00 & 54.75 & 56.25 & 57.25 \\ \hline
uni + bi + tri + swn + socal + nrc & 54.50 & 58.25 & \textbf{59.00} & 55.00 & 55.00 & 56.75 & 57.25 \\ \hline
\end{tabular}
\captionsetup{justification=centering}
\caption{Accuracies of classifiers (English train - code-mixed test). Boldface: Best performance.}
\label{table4}
\end{table}

\textbf{Summary}: In exp2, we can see a significant downfall in the accuracies of the classifiers as compared to exp1 (from 84.45\% to 59.00\%). In this experiment, we can again see a significant rise in the accuracies after adding adverbs, verbs and nouns along with adjectives (average 3.67\%). Introduction of bigrams (fe03) shows very small improvement (average 0.35\%) and even smaller (average 0.07\%) or no improvement (LSVC) in case of trigrams (fe04). For linear models (i.e. LRC and SGDC), we can see a drop in the accuracy after introduction of bi-grams and trigrams [12]. Introduction of SWN again shows a bit improvement in the accuracy (average 2.14\%). Adding SOCAL (fe06) and NRC emotion lexicon (fe07) did not show much improvement. \\
\textbf{Classifiers}: We can see from Table~\ref{table4} that Na\"ive Bayes (MNB and BNB) performed better than the rest. Among these two, MNB (confusion matrix shown in Table 8) performed better with an accuracy of 59.00\% whereas BNB got 58.25\%. For linear models, both LRC and SGDC had varying changes in the accuracy as different features were introduced but at the end, for fe07, both of them got the same accuracy of 55.00\%. Among Support Vector Machines, NuSVC performed a bit better with an accuracy of 57.25\% while LSVC got 56.75\%. In this case (English train - code-mixed test), based on performance, we can infer that NB > SVM > LM. \\ \\
\textbf{Exp3}: Here, we did few more preprocessing steps 1) removal of emoticons and
hashtags. Next, we collected top 1000 unigrams (fe08) based on tf-idf. Unigrams
which were either English stop-words were not chosen. Also, Bengali unigrams (e.g. er, e, je) which does not carry any sentiment were ignored as well. After extracting
bigrams, bigrams containing $\geq$ 1 English stop words or Bengali non sentiment
words were removed [13]. For trigrams, similar experiment was done
except trigrams containing $\geq$ 2 English stop words or Bengali non-sentiment words
were removed [13]. For bigrams (fe09), top 200 were chosen and trigrams top 100
(fe10) were chosen based on tf-idf. They were used in a bag of words fashion.
Lexicons used for feature extraction from code-mixed training set were SWN
(fe11), SOCAL (fe12) and NRC emotion lexicon (fe13). For NRC emotion lexicon,
words without a polarity (i.e. positive 0, negative 0) were not used. Features used for
training are described in FSet2.  \\ \\
Feature set acronyms (FSet2): \\ 
\hspace*{0.5cm}fe08 $\rightarrow$ unigrams top 1000 \\
\hspace*{0.5cm}fe09 $\rightarrow$ fe08 + bigrams top 200 \\
\hspace*{0.5cm}fe10 $\rightarrow$ fe09 + trigrams top 100 \\
\hspace*{0.5cm}fe11 $\rightarrow$ fe10 + swn \\
\hspace*{0.5cm}fe12 $\rightarrow$ fe11 + socal \\
\hspace*{0.5cm}fe13 $\rightarrow$ fe12 + nrc
\begin{table}[H]
\begin{tabular}{|l|c|c|c|c|c|c|c|}
\hline
\multicolumn{1}{|c|}{\multirow{3}{*}{Feature Combinations}} & \multicolumn{7}{c|}{\textbf{Accuracy in \%}} \\ \cline{2-8} 
\multicolumn{1}{|c|}{} & \multicolumn{3}{c|}{NB} & \multicolumn{2}{c|}{LM} & \multicolumn{2}{c|}{SVM} \\ \cline{2-8} 
\multicolumn{1}{|c|}{} & GNB & BNB & MNB & LRC & SGDC & LSVC & NuSVC \\ \hline
uni & 51.50 & 53.50 & 56.00 & 55.25 & 53.75 & 59.00 & 58.75 \\ \hline
uni + bi & 52.50 & 54.25 & 56.75 & 56.00 & 54.75 & 60.25 & 60.00 \\ \hline
uni + bi + tri & {52.75} & {54.50} & {57.25} & {56.25} & {54.75} & {61.00} & {60.75} \\ \hline
uni + bi + tri + swn & {58.25} & {64.75} & {66.50} & {66.00} & {63.50} & {72.25} & {71.50} \\ \hline
uni + bi + tri + swn + socal & 58.75 & 65.50 & 67.00 & 66.25 & 64.25 & 72.50 & 71.75 \\ \hline
uni + bi + tri + swn + socal + nrc & 58.75 & 65.75 & 67.75 & 66.75 & 64.75 & \textbf{72.50} & 72.00 \\ \hline
\end{tabular}
\captionsetup{justification=centering}
\caption{Accuracies of classifiers (code-mixed train - code-mixed test). Boldface: Best performance shown in the experiment.}
\label{table5}
\end{table}
\textbf{Summary}: For unigrams (fe08), we can see quite a bit rise (average 2.14\%) in the
accuracy as compared to unigrams (fe02) shown in Table~\ref{table5}. Introduction of bigrams
(fe09) and trigrams (fe10) doesn’t improve the accuracy much. Interestingly, introduction
of SWN (fe11) shows a big improvement (average 9.35\%) in the accuracy
for all the classifiers. This is due to the fact that our code-mixed dataset contained a
lot of sentiment carrying English words. Also, this improvement is not shown in Table
4 in case of fe05 because fe05 only contains English features while fe11 contains
both English as well as Bengali features (fe08, fe09, fe10). Again, like exp1 and exp2, introducing SOCAL (fe12) and NRC emotion lexicon (fe13) shows slight improvement.\\
\textbf{Classifiers}: It can be seen from Table~\ref{table5} that support vector machines performed better
than the rest by a big margin. Among SVMs, LSVC (confusion matrix shown in Table~\ref{table8}) got the better score with an accuracy of 72.50\% (rise in 13.5\% accuracy from
exp2) while NuSVC got 72.00\%. LRC scored better than SGDC among linear models
with an accuracy of 66.75\% while SGDC got 64.75\%. Naïve Bayes didn’t disappoint
much either, MNB got the highest with an accuracy of 67.75\% followed by BNB with
an accuracy of 65.75\% and trailed by GNB with an accuracy of 58.75\%. In this experiment,
we can see a similar performance pattern with exp1. Here based on performance
we can infer that SVM > NB > LM.

\section{Error Analysis}
\label{sec6}
In this section, we discuss the drawbacks of our systems. We have done error analysis
for results on both English data as well as code-mixed data. Analysis on code-mixed data
is done more extensively as compared to English data since English data analysis has
been covered by a lot researchers before. The best performing systems are evaluated
with the help of confusion matrix along with important parameters like accuracy,
precision, recall, f1-score, g-measure and matthews correlation coefficient
(MCC). \\ \\
1. English Train \textbf{--} English Test: LSVC performed the best with an accuracy of 84.45\%
(shown in Table~\ref{table3}). Other parameter values are shown in Table~\ref{table6}. \\
2. English Train \textbf{--} Code-Mix Test: MNB performed the best with an accuracy of
59.00\% (shown in Table~\ref{table4}). Other parameter values are shown in Table~\ref{table7}. \\
3. Code-Mix Train \textbf{--} Code-Mix Test: LSVC performed the best with an accuracy of
72.50\% (shown in Table~\ref{table5}). Other parameter values are shown in Table~\ref{table8}.

\subsection{English Train -- English Test}
LSVC (SVM based) performed the best on English train - English test with an
accuracy of 84.45\% which can be considered to be a satisfactory performance.
NuSVC (SVM based) got an accuracy of 84.37\% and got the second position. From TP and TN we can see that the classifier overall is quite balanced as well.
\begin{table}[H]
\begin{tabular}{|c|c|c|c|c|c|}
\hline
\multicolumn{2}{|c|}{} & \multicolumn{2}{c|}{Negative} & \multicolumn{2}{c|}{Positive} \\ \hline
\multicolumn{2}{|c|}{\textit{Negative}} & \multicolumn{2}{c|}{TN : 1036} & \multicolumn{2}{c|}{FP : 164} \\ \hline
\multicolumn{2}{|c|}{\textit{Positive}} & \multicolumn{2}{c|}{FN : 209} & \multicolumn{2}{c|}{TP : 991} \\ \hline
\textbf{Accuracy} & \textbf{Precision} & \textbf{Recall} & \textbf{F1 Score} & \textbf{G Measure} & \textbf{MCC} \\ \hline
84.45\% & 85.80\% & 82.58\% & 84.16\% & 84.17\% & 5.79\% \\ \hline
\end{tabular}
\centering
\caption{ LSVC trained on fe07 and tested on English testing dataset.}
\label{table6}
\end{table}

\subsection{English Train - Code-Mixed Test}
MNB (NB based) performed the best on English train - code-mixed test. Confusion
matrix along with values of some important parameters are shown in Table~\ref{table7}.
\begin{table}[H]
\begin{tabular}{|c|c|c|c|c|c|}
\hline
\multicolumn{2}{|c|}{} & \multicolumn{2}{c|}{Negative} & \multicolumn{2}{c|}{Positive} \\ \hline
\multicolumn{2}{|c|}{\textit{Negative}} & \multicolumn{2}{c|}{TN : 114} & \multicolumn{2}{c|}{FP : 86} \\ \hline
\multicolumn{2}{|c|}{\textit{Positive}} & \multicolumn{2}{c|}{FN : 78} & \multicolumn{2}{c|}{TP : 122} \\ \hline
\textbf{Accuracy} & \textbf{Precision} & \textbf{Recall} & \textbf{F1 Score} & \textbf{G Measure} & \textbf{MCC} \\ \hline
59.00\% & 58.65\% & 61.00\% & 59.80\% & 59.81\% & 2.57\% \\ \hline
\end{tabular}
\centering
\caption{MNB trained on fe07 and tested on code-mixed testing dataset.}
\label{table7}
\end{table}
The challenges faced on code-mixed data include old challenges known for English
data as well as new challenges for code-mixing property. Since the entire training set
was in English, classifiers were unable to identify code-mixed words which are playing
an important role in the sentence for imparting a sentiment (Sen 1). Majority of the
mis-classifications are due to this reason. \\ \\
\textbf{Sen 1}. (Movie)\textsubscript{EN} (tar)\textsubscript{BN} (print quality)\textsubscript{EN} (eto kharap je)\textsubscript{BN} (patience)\textsubscript{EN}
(chilo na purota dekhar)\textsubscript{BN} . (clf - manual: neg, exp2: pos) \\
\underline{Translation}: The print quality was so bad that I did not have the patience to
watch the whole movie . \\
Reason: Due to the fact that the system could not identify the word \textit{kharap}. The word
\textit{kharap}, meaning \textit{bad} is a pretty common word used in Bengali which was
unidentified by the classifier. Also, the word \textit{patience} is generally associated with
positive sentiment. \\ \\
\textbf{Sen 2}. (Finally)\textsubscript{EN} (bohudin por shobkichu bad diye tana ekta)\textsubscript{BN} (series)\textsubscript{EN}
(dekhte parlam)\textsubscript{BN} . (clf - manual: pos, exp2: neg) \\
\underline{Translation}: Finally after a lot of days I could watch a series leaving
everything aside . \\
Reason: Due to the fact that classifier thought the word \textit{bad} to be an English word
though in the given sentence, \textit{bad} is used as a Bengali word meaning aside. \\ \\
\textbf{Sen 3}. (Starting)\textsubscript{EN} (er diker)\textsubscript{BN} (portion)\textsubscript{EN} (tah kemon jeno)\textsubscript{BN}
(hollow)\textsubscript{EN} (lagchilo)\textsubscript{BN} (but)\textsubscript{EN} (shesher)\textsubscript{BN} (portion)\textsubscript{EN} (ta asadharon)\textsubscript{BN} . (clf - manual: pos, exp2: neg) \\
\underline{Translation}: Starting portion kind of felt hollow but the ending was awesome . \\
Reason: Probably, due to the contribution of the English word \textit{hollow}, the system
couldn’t identify the word \textit{asadharon} meaning awesome which has a stronger positive
sentiment value compared to \textit{hollow} which has a weaker negative sentiment value. \\ \\
\textbf{Sen 4}. (Jodio)\textsubscript{BN} (actors)\textsubscript{EN} (der)\textsubscript{BN} (performance was satisfactory)\textsubscript{EN} ,
(ami)\textsubscript{BN} (2/10)\textsubscript{EN} (er besi debo na)\textsubscript{BN} . (clf - manual: neg, exp2: pos) \\
\underline{Translation}: Even though the actors performance was satisfactory, I
wouldn’t rate it more than 2/10 . \\
Reason: The error in classification of Sen 4 is not code-mixed specific. Here the possible reason
for classifying it as positive by the system is due to the word \textit{satisfactorily}
even
though 2/10 is clearly a poor rating i.e. negative (similar to Sen 3). 

\subsection{Code-Mixed Train -- Code-Mixed Test}
Like English train - English test, LSVC (SVM based) performed the best on code-mixed
train - code-mixed Test. Examples of some snippets from code-mixed testing data
that were misclassified in exp2 but correctly classified in exp3. 
\begin{table}[H]
\begin{tabular}{|c|c|c|c|c|c|}
\hline
\multicolumn{2}{|c|}{} & \multicolumn{2}{c|}{Negative} & \multicolumn{2}{c|}{Positive} \\ \hline
\multicolumn{2}{|c|}{\textit{Negative}} & \multicolumn{2}{c|}{TN : 141} & \multicolumn{2}{c|}{FP : 59} \\ \hline
\multicolumn{2}{|c|}{\textit{Positive}} & \multicolumn{2}{c|}{FN : 51} & \multicolumn{2}{c|}{TP : 149} \\ \hline
\textbf{Accuracy} & \textbf{Precision} & \textbf{Recall} & \textbf{F1 Score} & \textbf{G Measure} & \textbf{MCC} \\ \hline
72.50\% & 71.63\% & 74.50\% & 73.03\% & 73.05\% & 6.98\% \\ \hline
\end{tabular}
\centering
\captionsetup{justification=centering}
\caption{Results of LSVC trained on fe13 and tested on code-mixed testing dataset.}
\label{table8}
\end{table}
\textbf{Sen 5}. (Erokom ekta baje)\textsubscript{BN} (cinema)\textsubscript{EN} (korar por kono)\textsubscript{BN} (excuse)\textsubscript{EN}
(e)\textsubscript{BN} (public accept)\textsubscript{EN} (korbe na)\textsubscript{BN} , (and)\textsubscript{EN} (na korar e kotha)\textsubscript{BN} . (clf
- manual: neg, exp2: pos, exp3: neg) \\
\underline{Translation}: The public won’t accept any kind of excuse after making such a
bad cinema , and they shouldn’t either . \\
Reason: The word \textit{baje} means \textit{bad}. Similar to \textit{bad} in English, \textit{baje} is a very common
word used in Bengali to criticize something or to describe something as not good. \\ \\ \\
\textbf{Sen 6}. (Les Miserables)\textsubscript{EN} (er plotter thekeo beshi bhalo laglo er)\textsubscript{BN}
(drama)\textsubscript{EN} (ar)\textsubscript{BN} (musicals)\textsubscript{EN} (gulo)\textsubscript{BN} . (clf - manual: pos, exp2: neg,
exp3: pos) \\
\underline{Translation}: In Les Miserables I like the drama and musicals more than the
plot.\\
Reason: The word \textit{bhalo} means good. Similar to good in English, \textit{bhalo} is a very
common word used in Bengali to appreciate something. Examples of some snippets
from code-mixed testing data that were misclassified in exp3 as well. \\ \\
\textbf{Sen 7}. (Jemon)\textsubscript{BN} (story)\textsubscript{EN} (temn)\textsubscript{BN} (cast)\textsubscript{EN} , (ghyam)\textsubscript{BN} (artwork)\textsubscript{EN}
. (Filmtar againste bolar moton kicchu pelame na)\textsubscript{BN} . \\
\underline{Translation}: Like story like cast, awesome artwork. I have nothing to say
against the film . (clf - manual: pos, exp3: neg) \\
Reason: Small code-mixed training data. The word ghyam meaning awesome or sometimes
fantastic was not identified which clearly carries a positive sentiment. Small
training set is a major issue for supervised classifiers. \\ \\
\textbf{Sen 8}. (Sherlock)\textsubscript{EN} (toh)\textsubscript{BN} (old times)\textsubscript{EN} (er motone ekhono ache)\textsubscript{BN} ,
(amar mote chilo na kharap)\textsubscript{BN} . (clf - manual: pos, exp3: neg) \\
\underline{Translation}: Sherlock is same as the old times, according to me it wasn’t bad. \\
\textbf{Sen 9}. (Starting)\textsubscript{EN} (tah orokom shundor kore pore je ki hoye gelo)\textsubscript{BN} ?! (Ekdome
bhalo na)\textsubscript{BN} . (clf - manual: neg, exp3: pos) \\
\underline{Translation}: With a starting as beautiful as that what happened later ?! Not
at all good . \\
Reason: In both Sen 8 and Sen 9, the error is due to the property of negation which is
quite commonly faced during sentiment analysis. Both the sentences use the negating
word na which in the first sentence can be translated to wasn’t and as not in the second
sentence. In both the translated sentences, the negating word, \textit{wasn’t} and \textit{not} is
following the sentiment carrying word, \textit{bad} and \textit{good} respectively. This is the
common pattern in English and most of the classifiers made in the past which take
into account the property of negation is based on this idea. The interesting thing is
that in Sen 8, the negating word \textit{na} follows the sentiment carrying word \textit{kharap} (English
pattern) while in Sen 9, the negating word \textit{na} is followed by the sentiment carrying
word \textit{bhalo}. \\ \\
\textbf{Sen 10}. (It seemed like director box office success)\textsubscript{EN} (er jonho jeno teno prokare
erokom ekta)\textsubscript{BN} (movie)\textsubscript{EN} (te)\textsubscript{BN} (musicals)\textsubscript{EN} (dhukiyeche)\textsubscript{BN} . (clf -
manual: neg, exp3: pos) \\
\underline{Translation}: It seemed like the director wanted to include musicals in such a
movie by any means possible just to get a Box Office success. \\
Reason: System couldn’t identify the idiom \textit{jeno teno prokare} which in the translated
sentence is by any means, which clearly portrays a negative sentiment in the sentence.
This Bengali idiom can be loosely matched with the English idiom \textit{by hook or by 
crook}. Some other errors that we noticed were related to context specific knowledge and
domain specific knowledge.

\subsection{Others}
Some interesting bigrams and trigrams were seen as well while experimentation. For
example, in the bigram (money holo)\textsubscript{BN} meaning \textit{thought}, a POS tagger will tag
the word \textit{money} as noun referring to the currency but clearly it was meant to be
something else. This can cause error specially if trained on a domain specific dataset
where \textit{money} leans towards a polarity. This can be seen in Sen 2 error. Quite a few
examples of such sort were found. For some words, though the root part was in
English, it couldn’t be identified due to the the addition of suffixes like \textit{ta}, \textit{e}, \textit{je}, \textit{ei}.
Some examples are time(ta), movie(ta), poor(e). A conventional stemmer doesn’t always work
on such cases. If this process is done properly it can prove to be quite useful for
sentiment extraction. 

\section{Conclusion \& Future Work}
\label{sec7}
In this paper, we have made an effort to perform binary sentiment analysis on English-Bengali code-mixed data using three types of supervised classifiers, Naïve Bayes (NB), Linear Models (LM) and Support Vector Machines (SVM). Classifiers trained on English features performed quite satisfactorily on English data (84.45\%) but poorly on code-mixed data (59.00\%). We used classifiers trained on code-mixed features as well and saw a big improvement in the accuracy (72.50\%). Similar resources were used in both the cases, i.e. n-grams and lexicons, thus showing that it is effective for code-mixed analysis as well. We can also conclude that SVM tends to perform well when the train and test data are of similar type (Table~\ref{table3}, Table~\ref{table5}) while NB tends to perform better when the train and test data are of slightly different type (Table~\ref{table4}). Also, including SWN in exp3 gave a boost in the accuracy showing that users tend to use English words carrying sentiment quite often in spite of writing in code-mixed. We have also analyzed misclassified data from both the experiments (exp2, exp3) and have found the probable reasons for it which can be fixed in the future for better results.      Our immediate goal is to collect more code-mixed data with varied topics (e.g. sports, politics, conversations, etc) and also from platforms other than Twitter and Facebook like blogs, websites, chat threads, etc. Collecting large quantities of data and cleaning them is challenge as well. By acquiring a large enough corpora with as much less noise as possible, it’ll be possible to build lexical tools for code-mixed data (e.g. Bengali SentiWordNet in Roman script). Smaller NLP tools, for example negation dictionary, stop-word dictionary, intensifier dictionary, idiom dictionary,  etc can also be made which can be quite useful for sentiment identification like it is in the case for English data. A method using language identification [14] followed by POS tagging [15] and then polarity detection can be tried as well. It’ll also be useful to explore in detail other machine learning algorithms (specially neural nets and modified decision trees) on such data and try out different types of cascading [16] and ensemble techniques, both known ones and modified ones with different combinations of parameters. Also, selection of features play a very important role in any supervised learning technique. Commonly used features like n-grams can be filtered with POS tags, stop words, etc. Features like capitalization, length of sentence, context features, quoted portion, emoticons, etc can be tried as well.

\end{document}